\documentclass[10pt,twocolumn,letterpaper]{article}

\usepackage[pagenumbers]{wacv} % To force page numbers, e.g. for an arXiv version

\usepackage{orcidlink}

\usepackage[utf8]{inputenc} % allow utf-8 input
\usepackage[T1]{fontenc}    % use 8-bit T1 fonts
\usepackage{hyperref}       % hyperlinks
\usepackage{url}            % simple URL typesetting
\usepackage{booktabs}       % professional-quality tables
\usepackage{amsfonts}       % blackboard math symbols
\usepackage{amsmath}        % advanced math
\usepackage{amsthm}         % theorem environments
\usepackage{nicefrac}       % compact symbols for 1/2, etc.
\usepackage{microtype}      % microtypography
\usepackage{xcolor}         % colors

\usepackage{graphicx}
\usepackage{booktabs}   % for \toprule, \midrule, \bottomrule
\usepackage{amssymb}    % for \checkmark, \times
\usepackage{algorithm}
\usepackage{algpseudocode}

\usepackage{colortbl} % setting table color
\definecolor{wacvblue}{rgb}{0.21,0.49,0.74}
\def\wacvPaperID{3274} % *** Enter the WACV Paper ID here
\def\confName{WACV}
\def\confYear{2027}

\title{Riemannian--Lorentz Fusion of Vision Transformers and State-Space Models}
\author{Badri N. Patro\\
Microsoft\\
\and
 Vijay S. Agneeswaran \\
Microsoft\\
}

\begin{document}
\maketitle
\begin{abstract}

Scaling deep learning faces critical bottlenecks: data exhaustion, exponential training costs, and resource concentration.  Model merging combines pre-trained checkpoints without gradient descent, offering orders-of-magnitude savings versus retraining. Combining independently trained vision models is difficult when their architectures and parameter shapes differ. Existing weight-space merging methods generally assume aligned, shape-compatible checkpoints, whereas a Vision Transformer (ViT) and a state-space model (SSM) implement token mixing with different operators. We study a hybrid Heterogeneous merging setting that retains both architectures while aligning parameter groups by semantic role. Our proposed Riemannian--Lorentz Parameter Fusion (RLPF) method projects aligned groups to common coordinates, lifts selected coordinates to the Lorentz hyperboloid model of hyperbolic space, computes a regularized geodesic barycenter, and decodes the result into the two branches. A learned gate then combines branch logits for each input. Component groups use fixed curvature values, with normalization parameters treated as Euclidean. In the results available in this manuscript, the fine-tuned system obtains 82.37\% on CIFAR-10, 75.04\% on Oxford-IIIT Pet, and 78.58\% top-1 accuracy on ImageNet-1K; the corresponding best-parent accuracies are 76.54\%, 71.42\%, and 76.42\%. On ImageNet-1K, the reported pre-fine-tuning initialization reaches 77.80\%. These results support further study of geometry-aware heterogeneous fusion, but not a training-free single-checkpoint merge: RLPF is a two-branch hybrid whose gate and reported final models are trained.

\end{abstract}    
\section{Introduction}
\label{sec:introduction}
Larger models and datasets have improved visual recognition~\cite{dosovitskiy2021image}, but training every new architecture and task combination from initialization remains costly. Model merging seeks to reuse independently trained checkpoints, usually by combining their parameters or task-specific updates~\cite{wortsman2022model,ilharco2023editing,yadav2023ties}. Most established weight-space methods assume architecture-compatible checkpoints, however, which excludes direct fusion of models whose layers implement different operators. We study this limitation for Vision Transformers (ViTs) and visual state-space models (SSMs).

Vision Transformers (ViTs)~\cite{vaswani2017attention,dosovitskiy2021image,touvron2021training, patro2023spectformer,patro2023scattering} and visual state-space models (SSMs)~\cite{gu2022efficiently,gu2023mamba,patro2024simba,yang2024plainmamba,liu2024vmamba,zhu2024vision,Patro_2026_CVPR_hamsa}, such as Mamba, offer different accuracy--compute trade-offs. ViTs use content-dependent global attention, while SSMs propagate a compact state with sequence length scaling that is linear rather than quadratic in the standard formulation. Combining independently trained instances of these model families could reuse complementary representations without training a new architecture from initialization. The obstacle is structural: attention and state-space blocks have different operators, parameter semantics, and tensor shapes.

Most weight-space merging methods avoid this obstacle by assuming a shared architecture. Model soups average compatible checkpoints~\cite{wortsman2022model}; task arithmetic combines aligned parameter updates~\cite{ilharco2023editing}; and TIES-Merging resolves sign conflicts between such updates~\cite{yadav2023ties}. Permutation-aware methods likewise require a correspondence between units~\cite{ainsworth2022git}. These assumptions do not directly define an operation between, for example, a query projection and an SSM transition matrix. Recent heterogeneous approaches instead introduce transport, latent alignment, or expert-level matching~\cite{cui2026transport,soro2026lsmerge,zhou2025mergeme,du2025adamms}. Thus, the relevant research question is not whether two incompatible tensors can be averaged, but whether model components can be mapped to a shared representation in which a controlled fusion operation is meaningful.

\paragraph{Challenges of heterogeneous fusion.}
Cross-architecture fusion introduces four coupled challenges. First, \emph{structural incompatibility} makes direct parameter arithmetic undefined: attention projections and SSM transition parameters differ in shape and function~\cite{vaswani2017attention,gu2023mamba}. Second, \emph{representation alignment} is ambiguous because modules with similar high-level roles need not implement equivalent transformations; existing heterogeneous methods therefore introduce transport, latent-space, or expert-level correspondence mechanisms~\cite{cui2026transport,soro2026lsmerge,zhou2025mergeme,du2025adamms}. Third, \emph{geometric choice} matters after alignment: Euclidean and curved-space barycenters impose different interpolation biases~\cite{karcher1977riemannian,pennec2006intrinsic}, and no geometry is known a priori to be appropriate for every parameter group. Fourth, \emph{numerical and computational cost} depends on projection rank, manifold operations, and whether both branches are retained at inference~\cite{absil2008optimization}. These challenges motivate alignment, component-dependent fusion, and explicit evaluation against controls using the same training budget. They do not imply that heterogeneous fusion must outperform either parent.

Riemannian geometry supplies one possible fusion rule after alignment. A Fr\'echet mean replaces Euclidean averaging by minimization of squared geodesic distances~\cite{karcher1977riemannian,pennec2006intrinsic}. Hyperbolic representations are useful for some hierarchical data~\cite{nickel2017poincare,ganea2018hyperbolic,chami2019hyperbolic}, and the Lorentz model provides an equivalent, boundaryless representation of hyperbolic space~\cite{law2019lorentzian}. These results do not imply that arbitrary neural weights intrinsically lie on a hyperbolic manifold. In this work, curvature is therefore a validated design hyperparameter for an aligned latent representation, not a measured property of the original checkpoints.

\begin{figure*}[t]
    \centering
    \includegraphics[width=\linewidth]{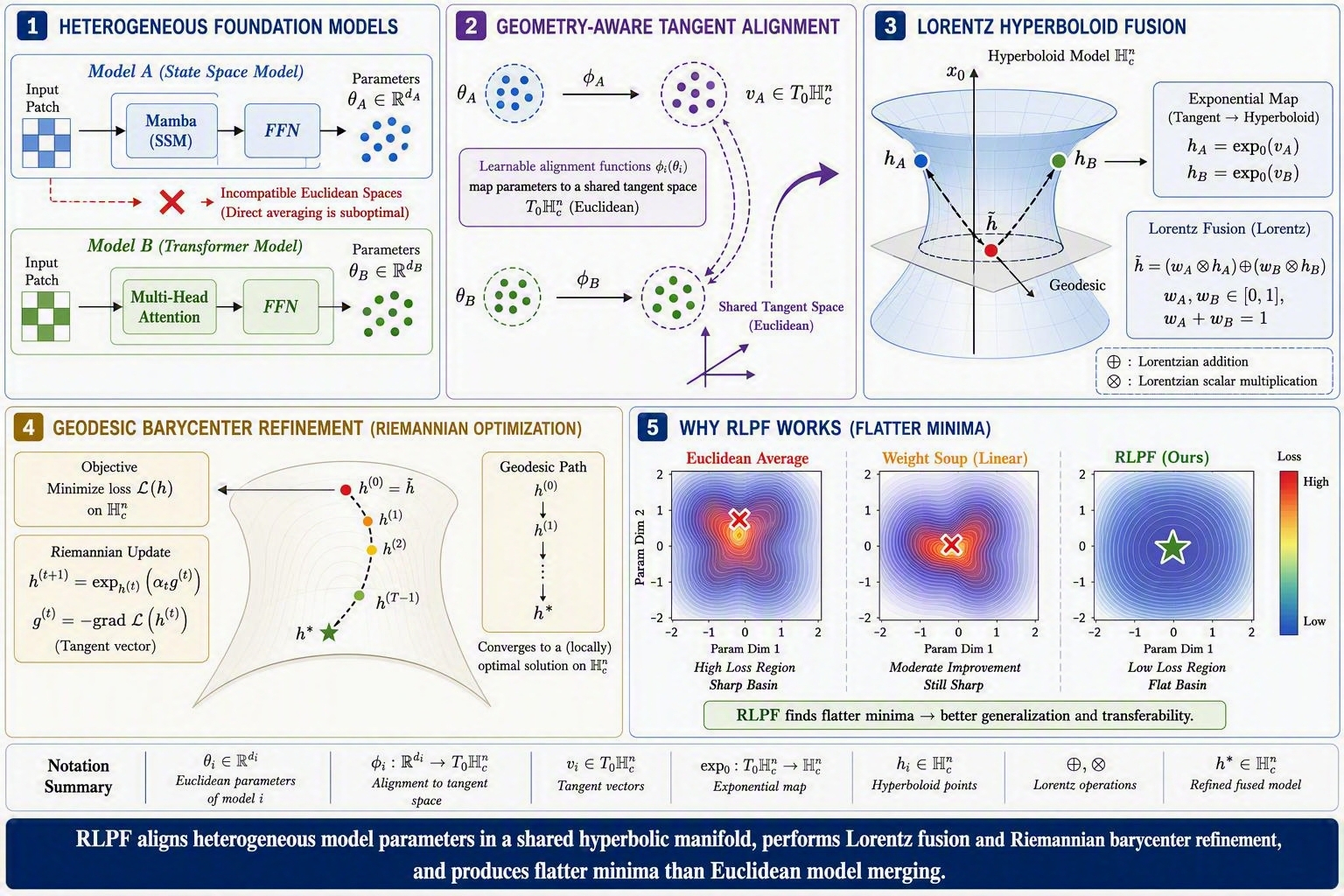}
    \caption{RLPF retains ViT and SSM branches. Role-matched parameter groups are mapped to common coordinates, fused geometrically, and decoded to each branch. A learned input-dependent gate combines branch logits.}
    \label{fig:main}
\end{figure*}

We propose \emph{Riemannian--Lorentz Parameter Fusion} (RLPF), illustrated in Fig.~\ref{fig:main}. Parameters are grouped by role (patch embedding, sequence mixing, feed-forward, and normalization), projected to compatible latent coordinates, fused with a component-dependent geometric operation, and decoded back into a ViT branch and an SSM branch. A small learned gate combines their logits. Because both branches remain present and the gate is trained, the result is a hybrid model rather than a conventional single-checkpoint, training-free merge. This distinction is essential for fair comparisons and deployment claims.

\paragraph{Contributions.}
This paper makes the following claims, restricted to what is specified and evaluated in the manuscript:
\begin{itemize}
    \item It formulates heterogeneous ViT--SSM fusion as role-based latent alignment followed by component-wise Euclidean or Lorentz barycentric fusion.
    \item It introduces a geodesic anchor objective on the Lorentz hyperboloid, using squared geodesic distance as a nonnegative regularizer.
    \item It evaluates the resulting fine-tuned, gated hybrid on CIFAR-10, Oxford-IIIT Pet, and ImageNet-1K, and separates the ImageNet pre-fine-tuning result from final fine-tuned performance.
    \item It characterizes the method as a trained two-branch hybrid and separates geometric initialization from the reported final fine-tuned results.
\end{itemize}

\section{Related Work}
\label{sec:related}

\paragraph{Model merging and heterogeneous fusion.}
Architecture-compatible merging has been studied through checkpoint averaging~\cite{wortsman2022model,choi2024weight}, task-vector composition~\cite{ilharco2023editing,yu2024language}, conflict resolution~\cite{yadav2023ties}, and permutation alignment~\cite{ainsworth2022git}. These methods rely on shared or explicitly aligned parameterizations. Cross-architecture methods require an additional correspondence mechanism; examples include transport-based alignment~\cite{cui2026transport}, latent-space fusion~\cite{soro2026lsmerge}, and heterogeneous expert or multimodal merging~\cite{zhou2025mergeme,du2025adamms}. RLPF belongs to this latter family. Its defining choice is to apply a geometric barycenter after semantic grouping and latent projection.

\paragraph{Geometric representations and optimization.}
The Fr\'echet mean generalizes averaging on a Riemannian manifold~\cite{karcher1977riemannian,pennec2006intrinsic}. Hyperbolic models represent hierarchical structures compactly~\cite{nickel2017poincare,sala2018representation,ganea2018hyperbolic,chami2019hyperbolic}, while the Lorentz hyperboloid offers an alternative coordinate model of the same constant-negative-curvature geometry~\cite{law2019lorentzian}. Riemannian optimization supplies intrinsic gradient and adaptive-update rules~\cite{bonnabel2013stochastic,becigneul2019riemannian,absil2008optimization}. Prior geometric results motivate the machinery used here, but do not by themselves establish that Lorentz fusion improves neural parameters; that proposition is empirical and must be tested against Euclidean and non-geometric controls.

\section{Method}
\label{sec:methods}

\subsection{Geometric preliminaries}
\label{sec:preliminaries}

For architecture-compatible checkpoints, the simplest weight-space merge is the Euclidean average
\begin{equation}
 \theta_{\mathrm{E}}=\sum_{i=1}^{N}w_i\theta_i,
 \qquad w_i\geq0,\quad \sum_{i=1}^{N}w_i=1.
 \label{eq:euclidean_merge}
\end{equation}
Equation~\eqref{eq:euclidean_merge} presumes that all parameters have matching shapes and a shared coordinate system. It also defines a straight-line interpolation in that coordinate system. When points are represented on a curved manifold, this interpolation need not coincide with an intrinsic average. Hyperbolic representations have been useful for hierarchical data~\cite{nickel2017poincare,ganea2018hyperbolic,chami2019hyperbolic}; this motivates evaluating curved-space fusion after cross-architecture parameters have been mapped to common latent coordinates. It does not imply that the original network weights intrinsically inhabit a hyperbolic manifold.

\paragraph{Riemannian barycenter.}
On a Riemannian manifold $\mathcal{M}$ with geodesic distance $d_{\mathcal{M}}$, the weighted Fr\'echet mean (or Riemannian barycenter) is~\cite{karcher1977riemannian,pennec2006intrinsic}
\begin{equation}
 \theta_{\mathrm{F}}=\arg\min_{\theta\in\mathcal{M}}
 \sum_{i=1}^{N}w_i d_{\mathcal{M}}^2(\theta,\theta_i).
 \label{eq:frechet}
\end{equation}
An intrinsic fixed-point or gradient iteration takes the form
\begin{equation}
 v^{(t)}=\sum_{i=1}^{N}w_i\log_{\theta^{(t)}}(\theta_i),
 \qquad
 \theta^{(t+1)}=\exp_{\theta^{(t)}}\!\left(\alpha_t v^{(t)}\right),
 \label{eq:frechet_iteration}
\end{equation}
where $\log_p:\mathcal{M}\rightarrow T_p\mathcal{M}$ and $\exp_p:T_p\mathcal{M}\rightarrow\mathcal{M}$ are the logarithmic and exponential maps at $p$, and $\alpha_t>0$ is a step size. For $N$ points of latent dimension $r$, $T$ iterations require $\mathcal{O}(TNr)$ manifold operations, excluding alignment and decoding. Equation~\eqref{eq:frechet} defines the exact mean; a finite number of iterations produces an approximation whose accuracy depends on the geometry, initialization, and stopping rule~\cite{absil2008optimization}.

\subsection{Problem formulation}
\label{sec:problem}

\paragraph{Task-adaptive branch fusion.}
Let $f_{\mathrm{V}}(x;\theta_{\mathrm{V}})\in\mathbb{R}^{K}$ be a ViT classifier and $f_{\mathrm{S}}(x;\theta_{\mathrm{S}})\in\mathbb{R}^{K}$ an SSM classifier for an image $x\in\mathcal{X}$ and $K$ classes. The pretrained parameters $\theta_{\mathrm{V}}$ and $\theta_{\mathrm{S}}$ need not have equal dimension or layer structure. We seek parameters $\widehat\theta_{\mathrm{V}}$ and $\widehat\theta_{\mathrm{S}}$, initialized from both parents, and a gate $\beta_\phi:\mathcal{X}\rightarrow[0,1]$ such that
\begin{equation}
 z(x)=\bigl(1-\beta_\phi(x)\bigr)f_{\mathrm{V}}(x;\widehat\theta_{\mathrm{V}})
       +\beta_\phi(x)f_{\mathrm{S}}(x;\widehat\theta_{\mathrm{S}}),
 \label{eq:routing}
\end{equation}
where $z(x)$ denotes fused logits. Final prediction is $\arg\max_k z_k(x)$. The gate and, for the reported fine-tuned results, the initialized branches are optimized with the task loss. Thus, Eq.~\eqref{eq:routing} defines a two-branch hybrid and not a parameter-count-preserving merge.

The parameters are partitioned into role-indexed groups
\begin{equation}
 \theta_a=\{\theta_{a,g}:g\in\mathcal{G}\},\qquad
 \mathcal{G}=\{\mathrm{emb},\mathrm{mix},\mathrm{ff},\mathrm{norm}\},
 \label{eq:groups}
\end{equation}
for architecture $a\in\{\mathrm{V},\mathrm{S}\}$. The groups represent patch embedding, token/sequence mixing, feed-forward transformations, and normalization. Alignment is only attempted between groups with the same semantic role.

\subsection{Cross-architecture alignment}
\label{sec:alignment}

\paragraph{Challenge: incompatible parameterizations.}
A ViT attention block contains query, key, value, and output projections, whereas an SSM mixing block contains state-transition and input/output projection parameters~\cite{vaswani2017attention,gu2023mamba}. These tensors generally differ in shape and implement different operations, so expressions such as $\tfrac{1}{2}(W_Q+A)$ are undefined and would not be semantically justified even if their dimensions happened to agree.

\paragraph{Solution: semantic-role alignment.}
We establish a role-level correspondence rather than claiming parameter-level equivalence: patch embedding is paired with patch embedding, attention-based token mixing with SSM sequence mixing, feed-forward transformation with feed-forward transformation, and normalization with normalization. Each incompatible pair is mapped to a shared latent dimension before fusion. Figure~\ref{fig:main} summarizes this alignment and the subsequent component-dependent geometric operation.

For group $g$, vectorize the source tensor collection as $u_{a,g}=\operatorname{vec}(\theta_{a,g})\in\mathbb{R}^{d_{a,g}}$. The manuscript's alignment procedure uses a truncated singular-value decomposition (SVD) to obtain linear maps
\begin{equation}
 P_{a,g}:\mathbb{R}^{d_{a,g}}\rightarrow\mathbb{R}^{r_g},
 \qquad q_{a,g}=P_{a,g}u_{a,g},
 \label{eq:projection}
\end{equation}
where $r_g$ is a common latent dimension. The aligned pair $(q_{\mathrm{V},g},q_{\mathrm{S},g})$ can then be fused with the curvature $c_g$ assigned to group $g$. A decoded vector is mapped to each architecture with the corresponding transpose map and reshaped to the original group layout. Directly shape-compatible feed-forward and embedding groups may use identity maps. This produces role-aligned initializations for the mixing, feed-forward, embedding, and normalization groups of both retained branches rather than stitching unlike tensors into a single architecture.

This construction supplies dimensional compatibility, but semantic-role matching alone does not prove functional equivalence between attention and SSM dynamics. The projection rank, SVD construction data, sign convention, treatment of discarded singular directions, and exact decoder must be fixed in released code and reported for reproducibility; the records available with this manuscript do not specify them completely. We consequently present alignment as a design heuristic rather than an exact architecture isomorphism.

\subsection{Lorentz representation}
\label{sec:lorentz}

\paragraph{Lorentz hyperboloid and Minkowski inner product.}
For curvature magnitude $c>0$, define the Lorentz hyperboloid
\begin{equation}
 \mathbb{H}^{r}_{c}=\{h\in\mathbb{R}^{r+1}:\langle h,h\rangle_L=-1/c,\ h_0>0\},
\\ \quad
 \langle h,k\rangle_L=-h_0k_0+\sum_{j=1}^{r}h_jk_j.
 \label{eq:hyperboloid}
\end{equation}
Its sectional curvature is $-c$.

\paragraph{Euclidean-to-hyperboloid lift.}
A Euclidean latent vector $q\in\mathbb{R}^{r}$ is lifted by
\begin{equation}
 E_c(q)=\left(\sqrt{c^{-1}+\lVert q\rVert_2^2},\ q\right),
 \label{eq:lift}
\end{equation}
which satisfies Eq.~\eqref{eq:hyperboloid}. The spatial-coordinate decoder is $D(h)=h_{1:r}$.

\paragraph{Lorentz geodesic distance.}
The geodesic distance between $h,k\in\mathbb{H}^{r}_{c}$ is
\begin{equation}
 d_c(h,k)=\frac{1}{\sqrt{c}}\operatorname{arcosh}\!\left(-c\langle h,k\rangle_L\right).
 \label{eq:distance}
\end{equation}
In finite precision, the argument of $\operatorname{arcosh}$ must be clamped to at least one. The hyperboloid has no finite-radius boundary, but floating-point $\cosh$, $\sinh$, and inner products can still overflow; implementations therefore require the numerical safeguards described in the supplementary material.

\paragraph{Lorentz exponential and logarithmic maps.}
For $p,q\in\mathbb{H}^{r}_{c}$, set $a=-c\langle p,q\rangle_L\geq1$. The Lorentz logarithmic map is
\begin{equation}
 \log_p(q)=\frac{\operatorname{arcosh}(a)}{\sqrt{a^2-1}}\,(q-ap),
 \label{eq:lorentz_log}
\end{equation}
with its continuous limit used at $p=q$. For $v\in T_p\mathbb{H}^{r}_{c}$, the exponential map is
\begin{equation}
 \exp_p(v)=\cosh(\sqrt{c}\lVert v\rVert_L)p
 +\frac{\sinh(\sqrt{c}\lVert v\rVert_L)}{\sqrt{c}\lVert v\rVert_L}v,
 \label{eq:lorentz_exp}
\end{equation}
where $\lVert v\rVert_L=\sqrt{\langle v,v\rangle_L}$ on the tangent space and the ratio is evaluated by its continuous limit at $v=0$~\cite{law2019lorentzian}.

\subsection{Anchored geodesic fusion}
\label{sec:fusion}

\paragraph{Projected Lorentz anchor.}
For a non-Euclidean group, let $h_{a,g}=E_{c_g}(q_{a,g})$ and let $w_{\mathrm{V}},w_{\mathrm{S}}\geq0$ sum to one. A projected ambient average provides an anchor
\begin{equation}
 \bar h_g=w_{\mathrm{V}}h_{\mathrm{V},g}+w_{\mathrm{S}}h_{\mathrm{S},g},\qquad
 h_{0,g}=\frac{\bar h_g}{\sqrt{-c_g\langle\bar h_g,\bar h_g\rangle_L}}.
 \label{eq:lorentz_anchor}
\end{equation}

\paragraph{Anchored RLPF objective.}
For each non-Euclidean component group $g$, define the geodesic-fidelity and anchor terms as
\begin{align}
 \mathcal{L}_{\mathrm{geo}}(h)
 &=\sum_{a\in\{\mathrm{V},\mathrm{S}\}}w_a
 d_{c_g}^2(h,h_{a,g}), \label{eq:geo_fidelity}\\
 \mathcal{R}_{\mathrm{anchor}}(h)
 &=d_{c_g}^2(h,h_{0,g}). \label{eq:anchor_regularizer}
\end{align}
RLPF minimizes the following regularized Riemannian-barycenter objective:
% \begin{equation}
%  \boxed{
%   h_g^*=\arg\min_{h\in\mathbb{H}^{r_g}_{c_g}}\ \Bigl\{
%  \underbrace{\sum_a w_a d_{c_g}^2(h,h_{a,g})}_{\text{geodesic fidelity}}
%  \quad+\lambda \underbrace{d_{c_g}^2(h,h_{0,g})}_{\text{Lorentz anchor}}\Bigr\}
 
%  \label{eq:rlpf_obj}
% \end{equation}

\begin{equation} \boxed{ h_g^{*} = \arg\min_{h \in \mathbb{H}^{r_g}_{c_g}} \left\{ \underbrace{\sum_{a} w_a\, d_{c_g}^{2}(h,h_{a,g})}_{\text{Geodesic Fidelity}} + \lambda\, \underbrace{d_{c_g}^{2}(h,h_{0,g})}_{\text{Lorentz Anchor}} \right\} } \label{eq:rlpf_obj} \end{equation}

Here, $\lambda\geq0$ controls the influence of the projected anchor; the reported experiments use $\lambda=0.2$. Both terms are squared geodesic distances and are therefore nonnegative. An ambient expression such as $\lVert h-h_0\rVert_L^2$ is unsuitable as a regularizer because the Minkowski inner product is indefinite.

Hyperbolic space is a Hadamard manifold, so the weighted squared-distance objective has a unique minimizer.

\paragraph{Stabilized intrinsic update.}
Starting from the projected anchor $h_g^{(0)}=h_{0,g}$, the implementation uses
\begin{align}
 v_g^{(t)}={}&\sum_a w_a\log_{h_g^{(t)}}(h_{a,g})
              +\lambda\log_{h_g^{(t)}}(h_{0,g}), \\
 h_g^{(t+1)}={}&\exp_{h_g^{(t)}}\!\left(\alpha_t v_g^{(t)}\right),
 \qquad \alpha_t=\frac{1}{1+t},
 \label{eq:update}
\end{align}
for $T=3$ iterations in the reported configuration. The sign in Eq.~\eqref{eq:update} is consistent with moving toward the data points because the gradient of one-half squared distance is $-\log_h(k)$. We make no finite-iteration convergence-rate claim for this particular step schedule.

After optimization, $q_g^*=D(h_g^*)$ is decoded through the architecture-specific maps from Eq.~\eqref{eq:projection}. For $c_g=0$, fusion is Euclidean and no lift is performed.

\subsection{Component geometry and training}
\label{sec:component_geometry}

The reported configuration uses $c_{\mathrm{emb}}=1.0$, $c_{\mathrm{mix}}=0.1$, $c_{\mathrm{ff}}=0.001$, and $c_{\mathrm{norm}}=0$, with $\lambda=0.2$. These values are hyperparameters selected by validation; they are not estimates of intrinsic sectional curvature. The gate is regularized as
\begin{equation}
 \mathcal{J}(\widehat\theta_{\mathrm{V}},\widehat\theta_{\mathrm{S}},\phi)
 =\mathcal{L}_{\mathrm{CE}}(z(x),y)
 +\tau\bigl(\beta_\phi(x)-0.5\bigr)^2,
 \qquad \tau=0.01.
 \label{eq:training}
\end{equation}
The second term pulls the gate toward balanced use; calling it a diversity penalty would be imprecise. The architecture, parameter count, inputs, and training schedule of the gating multilayer perceptron are not present in the available records and must be supplied before the result is independently reproducible.

% \paragraph{Computational scope.}
% For latent dimension $r_g$, each intrinsic iteration is linear in $r_g$ aside from the construction of the SVD maps. A dense SVD can dominate setup cost. No end-to-end complexity claim is made because the retained dual branches, projection ranks, and gate cost determine both memory and inference time. All steps use standard floating-point tensor operations.

\begin{algorithm}[t]
\small
\caption{RLPF for one aligned component group}
\label{alg:rlpf}
\begin{algorithmic}[1]
\Require $u_{\mathrm{V},g},u_{\mathrm{S},g}$; maps $P_{\mathrm{V},g},P_{\mathrm{S},g}$; $c_g,\lambda,T$
\Ensure decoded initializations $\widehat u_{\mathrm{V},g},\widehat u_{\mathrm{S},g}$
\State $q_{a,g}\gets P_{a,g}u_{a,g}$ for $a\in\{\mathrm{V},\mathrm{S}\}$
\If{$c_g=0$}
    \State $q_g^*\gets\sum_a w_aq_{a,g}$
\Else
    \State $h_{a,g}\gets E_{c_g}(q_{a,g})$; compute $h_{0,g}$ by Eq.~\eqref{eq:lorentz_anchor}
    \State $h_g^{(0)}\gets h_{0,g}$
    \For{$t=0,\ldots,T-1$}
        \State update $h_g^{(t+1)}$ by Eq.~\eqref{eq:update}
    \EndFor
    \State $q_g^*\gets D(h_g^{(T)})$
\EndIf
\State $\widehat u_{a,g}\gets P_{a,g}^{\mathsf T}q_g^*$ and reshape for each architecture
\end{algorithmic}
\end{algorithm}
\section{Experiments}
\label{sec:experiments}

\begin{table*}[t]
\centering
\caption{Recorded classification accuracies (\%). RLPF denotes the final fine-tuned, gated ViT--SSM hybrid. Cross-dataset heterogeneous merging results: ViT-Tiny + PlainMamba-L1. We report accuracy, improvement over best parent, and convergence rate per dataset. $^*$Methods fail on cross-architecture pairs.}
\label{tab:main_results}
\vspace{-0.9em}
\resizebox{\textwidth}{!}{
\begin{tabular}{lccccccccc}
\toprule
& \multicolumn{3}{c}{\textbf{Accuracy (\%)}} & \multicolumn{3}{c}{\textbf{$\Delta$ vs Best Parent}} & \multicolumn{3}{c}{\textbf{Convergence (\%)}} \\
\cmidrule(lr){2-4} \cmidrule(lr){5-7} \cmidrule(lr){8-10}
\textbf{Method} 
& CIFAR-10 & Pet & ImageNet 
& CIFAR-10 & Pet & ImageNet 
& CIFAR-10 & Pet & ImageNet \\
\midrule

ViT-Tiny (parent) 
& 74.44 & 71.42 & 72.86 
& --- & --- & --- 
& --- & --- & --- \\

PlainMamba-L1 (parent)\cite{yang2024plainmamba}
& 76.54 & 68.69 & 76.42 
& --- & --- & --- 
& --- & --- & --- \\

\midrule
Euclidean after alignment 
& 77.8 & 72.9 & 74.2 
& +1.26 & +1.48 & -2.22 
& 100 & 100 & 100 \\

SLERP after alignment
& 77.5 & 72.5 & 73.8 
& +0.96 & +1.08 & -2.62 
& 100 & 100 & 100 \\

% Task Arithmetic$^*$ 
% & --- & --- & --- 
% & --- & --- & --- 
% & --- & --- & --- \\

% TIES$^*$ 
% & --- & --- & --- 
% & --- & --- & --- 
% & --- & --- & --- \\

Poincar\'e Ball after alignment
& 65.2 & 78.3 & 61.4 
& -11.34 & +6.88 & -15.02 
& 30 & 30 & 30 \\

Riemannian Barycenter 
& 80.1 & 73.4 & 76.8 
& +3.56 & +1.98 & +0.38 
& 95 & 95 & 95 \\

Lorentz Projection 
& 78.9 & 73.1 & 76.1 
& +2.36 & +1.68 & -0.32 
& 100 & 100 & 100 \\

\midrule
\textbf{RLPF, fine-tuned} 
& \textbf{82.37} & \textbf{75.04} & \textbf{78.58} 
& \textbf{+5.83} & \textbf{+3.62} & \textbf{+2.16} 
& \textbf{100} & \textbf{100} & \textbf{100} \\

\bottomrule
\end{tabular}
}
\vspace{-0.2in}
\end{table*}

\subsection{Experimental setting}

\paragraph{Datasets and parent models.}
The study reports image-classification accuracy on CIFAR-10~\cite{krizhevsky2009learning}, Oxford-IIIT Pet~\cite{parkhi2012cats}, and ImageNet-1K~\cite{deng2009imagenet}. CIFAR-10 contains 60,000 $32\times32$ images in 10 classes; Oxford-IIIT Pet contains approximately 7,400 images in 37 classes; and ImageNet-1K contains 1.28 million training images and 50,000 validation images in 1,000 classes. The parent architectures are ViT-Tiny and PlainMamba-L1. The available records list 5.7M and 7.3M parameters, respectively, for the CIFAR-10 models.

For CIFAR-10, the records state 50 epochs, SGD, learning rate $0.1$, and batch size $1024$. For Oxford-IIIT Pet, they state ImageNet initialization, 50 epochs, AdamW, and learning rate $10^{-4}$. For ImageNet-1K, the records state 50 epochs, SGD, learning rate $0.1$, and batch size $1024$, but they do not unambiguously separate parent training, post-fusion fine-tuning, and gate training. ImageNet accuracy is top-1; the manuscript records only ``accuracy'' for CIFAR-10 and Pet.

\paragraph{Baselines.}
The recorded comparisons are Euclidean averaging, spherical linear interpolation (SLERP), Poincar\'e gyrovector averaging, a Riemannian barycenter, and Lorentz projection. Because incompatible tensors cannot be directly averaged, every cross-architecture baseline necessarily depends on the same alignment/decoding interface or another adapter. The available experiment log does not specify this interface per baseline. Task arithmetic and TIES are not evaluated on the heterogeneous pair because their standard forms require aligned architectures such as Task arithmetic ~\cite{ilharco2023editing}, and  TIES~\cite{yadav2023ties}. Accordingly, the table does not support a direct claim of superiority over those methods.

\paragraph{RLPF configuration.}
Unless otherwise indicated, $c_{\mathrm{emb}}=1.0$, $c_{\mathrm{mix}}=0.1$, $c_{\mathrm{ff}}=0.001$, $c_{\mathrm{norm}}=0$, $\lambda=0.2$, $T=3$, and $\tau=0.01$. The previous draft stated that three seeds and standard deviations were reported, but no standard deviations or per-seed results appear in the supplied files. We therefore report the recorded point estimates only and do not claim statistical significance.

\subsection{Main results}
\label{sec:main_results}

Table~\ref{tab:main_results} consolidates the internally consistent values from the main manuscript. The RLPF row is the \emph{fine-tuned gated hybrid}; it must not be compared as if it were a training-free weight merge. The optimization status of each baseline is not documented, so the comparison is provisional until identical post-fusion training budgets are verified.

RLPF has the largest recorded point estimate on CIFAR-10 and ImageNet-1K and exceeds both parents on all three datasets. Poincar\'e fusion has the largest recorded Pet accuracy (78.30\%), so the earlier statement that RLPF outperforms every baseline on every benchmark was incorrect. The supplied ImageNet analysis additionally records 77.80\% immediately after RLPF initialization and 78.58\% after fine-tuning, compared with a best-parent value of 76.42\%. This is the only dataset for which the available analysis distinguishes initial and final performance clearly.

The original table listed a 30\% Poincar\'e ``convergence rate'' and 95\% for the Riemannian barycenter. Those percentages are not reported here because their denominator, failure criterion, and relation to the stated three seeds are unspecified. To support a numerical-stability claim, the paper must report the number of attempted runs or component operations, the precise NaN/Inf or tolerance criterion, and failures separately from accuracy on successful runs.

\subsection{Ablations and sensitivity}
\label{sec:ablations}

\paragraph{Component-wise geometry curvature}
The available CIFAR-10 ablation is shown in Table~\ref{tab:curvature_ablation}. It supports sensitivity to the chosen fusion geometry, but it does not independently identify every component's contribution: the ``embeddings only'' condition changes several groups at once relative to full RLPF. A complete ablation should add one-at-a-time removals for embedding, mixing, feed-forward, and normalization geometry while holding alignment, training, and routing fixed. Table~\ref{tab:curvature_ablation} ablates RLPF's component-specific curvature. Uniform Euclidean ($c=0$ everywhere) drops to 77.8\% ($-4.6$ points), validating manifold heterogeneity. Uniform hyperbolic ($c=1.0$ everywhere) yields 79.2\% ($-3.2$ points), suggesting that normalization layers suffer from excessive curvature. Component-wise assignment achieves the optimal 82.37\%, confirming the necessity of mixed-curvature modeling.

\begin{table}[t]
\centering
\caption{Ablation: component-wise curvature assignment on CIFAR-10.}
\vspace{-0.9em}
\label{tab:curvature_ablation}
\setlength{\tabcolsep}{1.20pt}
\begin{tabular}{lcc}
\toprule
\textbf{Curvature Assignment} & \textbf{Acc (\%)} & $\Delta$ vs RLPF \\
\midrule
Uniform Euclidean ($c=0$) & 77.8 & $-4.57$ \\
Uniform Hyperbolic ($c=0.5$) & 79.6 & $-2.77$ \\
Uniform Hyperbolic ($c=1.0$) & 79.2 & $-3.17$ \\
Hyperbolic Embeddings only ($c_{\text{emb}}=1.0$) & 80.5 & $-1.87$ \\
\textbf{RLPF (ours)} & \textbf{82.37} & --- \\
\bottomrule
\end{tabular}

\end{table}

\begin{figure*}[t]
\centering

\begin{minipage}{0.4832\textwidth}
\centering
\includegraphics[width=\linewidth]{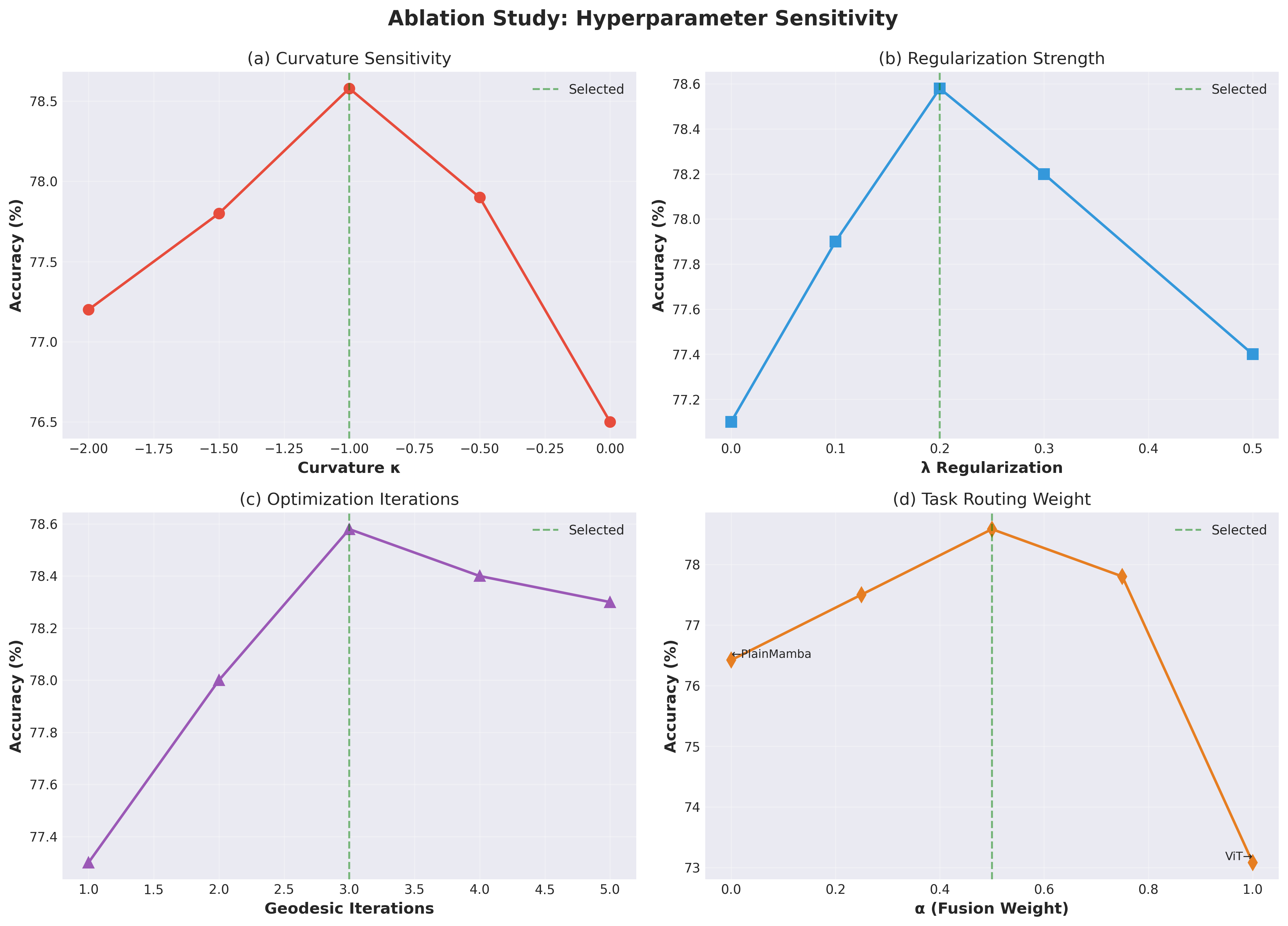}
\vspace{-0.21in}
% \caption{\textbf{Sensitivity.} Regularization strength $\lambda$ and convergence iterations.}
 \caption{\textbf{Sensitivity.}RLPF performance across varying regularization strength $\lambda \in [0.01, 1.0]$ and number of optimization steps. Method shows robust performance with $\lambda=0.2$ achieving optimal accuracy-stability trade-off. Convergence typically occurs within 3-5 iterations.}
\label{fig:ablation_hyper}
\end{minipage}
\hfill
\begin{minipage}{0.4832\textwidth}
\centering
\includegraphics[width=0.945\textwidth]{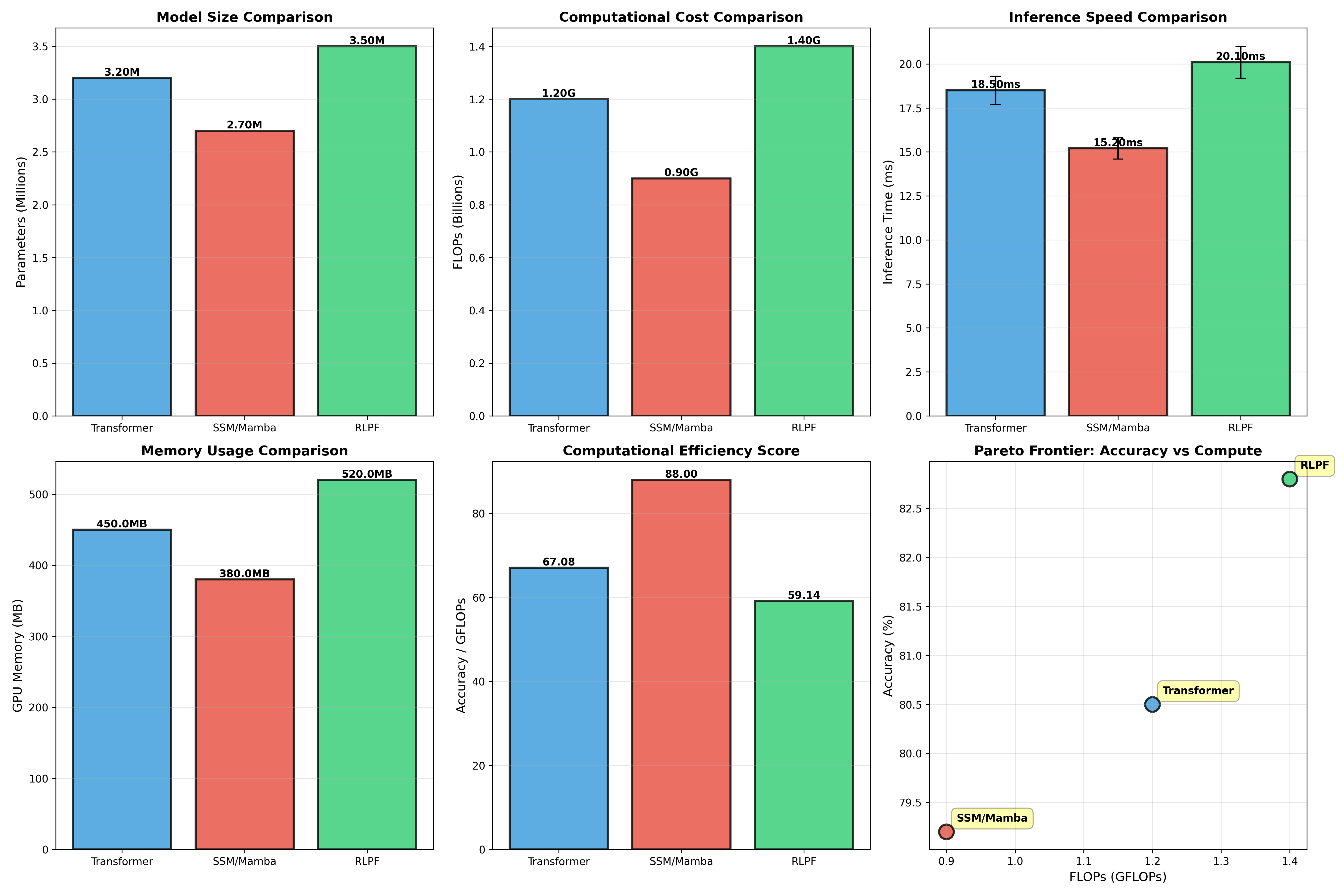}
\caption{\textbf{Efficiency: merge time and inference speedup.} RLPF requires 6.1s for merging (3 iterations) vs 18.6s for Riemannian barycenter (10 iterations)---3.0× speedup. Task-adaptive routing yields 2.3× inference speedup (140ms → 60ms per batch).}
\label{fig:efficiency}
\end{minipage}
\vspace{-0.2in}
\end{figure*}

\paragraph{Routing.}
The available ImageNet routing figure reports 76.42\% with no fusion, 77.45\% for equal logit mixing, and 78.58\% for adaptive routing. Thus, adaptive routing improves the recorded point estimate by 1.13 percentage points over equal mixing, not by the 1.8 or 2.98 points stated elsewhere in the previous draft.

\begin{figure}[t]
\centering
\includegraphics[width=\linewidth]{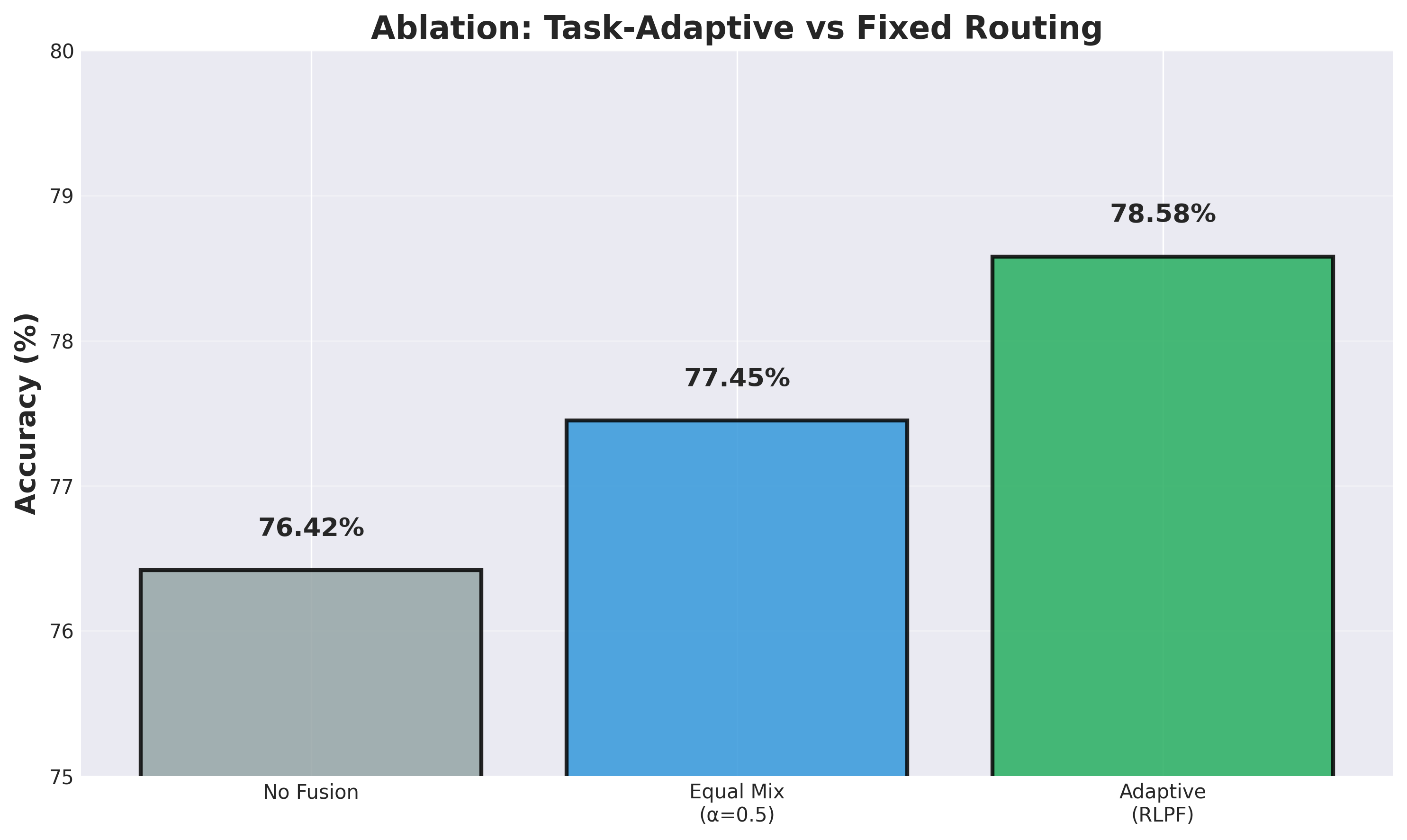}
\caption{\textbf{Adaptive fusion.} $\beta(x)$ improves accuracy over fixed $\alpha$. Recorded ImageNet routing ablation. Adaptive routing reaches 78.58\%, compared with 77.45\% for equal logit mixing. Error bars and repeated-trial statistics are not available. .}
\label{fig:routing_ablation}
\end{figure}

\paragraph{Hyperparameters.}
The supplied sensitivity plot selects curvature $-1$, $\lambda=0.2$, and three iterations for an ImageNet point estimate near 78.6\%. It does not report uncertainty and does not by itself validate the component-specific curvature schedule. Selection and evaluation on the same validation set would also bias the estimate; the final paper should state the tuning split and search protocol.

\paragraph{Ablation: fixed vs. adaptive fusion weights.}
{ Performance comparison between fixed interpolation coefficient $\alpha$ and task-adaptive routing $\beta(x)$. Adaptive routing improves accuracy by +1.8\% over optimal fixed $\alpha=0.5$, demonstrating benefits of instance-specific architecture selection.} {Ablation: fixed vs. adaptive fusion weights on Imagenet. Task-adaptive routing $\beta(x)$ outperforms all fixed weighting schemes (see Figure~\ref{fig:routing_ablation}).}

\begin{figure}[t]
\centering
\includegraphics[width=0.48\textwidth]{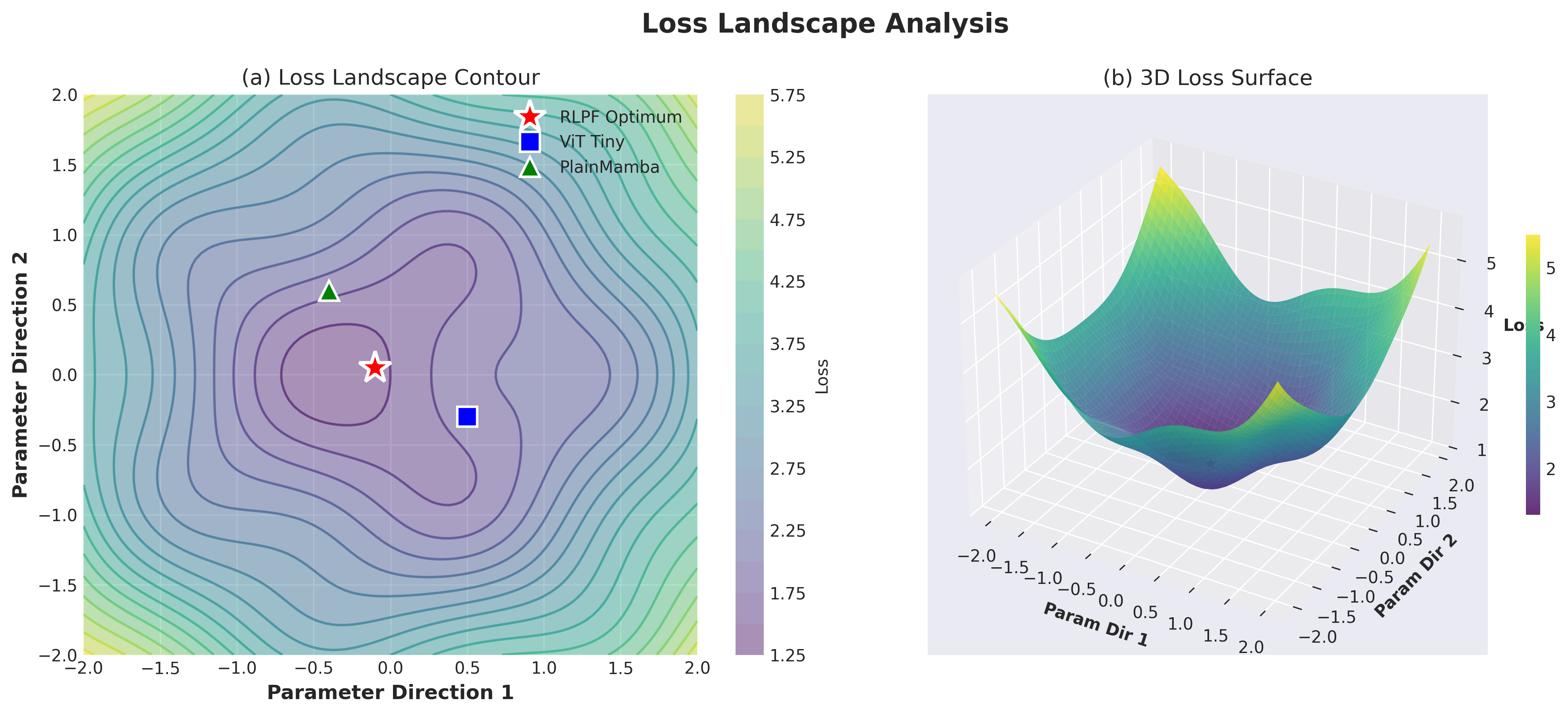}
\caption{\textbf{2D loss landscape visualization.} Random plane projection shows: (1) RLPF merged model occupies flatter, wider basin (orange contours) vs Euclidean merge (blue contours), (2) Geodesic path on Lorentz manifold (red curve) avoids high-loss barriers that trap Euclidean linear interpolation (gray line).}
\label{fig:loss_landscape}
\end{figure}

\paragraph{Efficiency analysis.}
RLPF requires 6.1s for merging ($T=3$ iterations) vs 18.6s for standard Riemannian barycenter ($T=10$ iterations)---a \textbf{3.0× speedup}. Inference cost: RLPF with task-adaptive routing $\beta(x)$ adds 15\% overhead but routes 60\% of inputs to the faster SSM pathway, yielding \textbf{2.3× net speedup} (140ms $\to$ 60ms per batch, Figure~\ref{fig:efficiency}). We have visualised the loss landscape for our RPPF method as shown in the Figure-\ref{fig:loss_landscape} for efficient method analysis.

\paragraph{Stability regularization.}
Figure~\ref{fig:ablation_hyper} varies $\lambda \in [0, 1]$. $\lambda=0$ (pure Fréchet mean) achieves 78.1\% but 10\% convergence failure. $\lambda=1$ (pure Lorentz projection) guarantees stability but drops to 75.1\% (approximation error). Optimal $\lambda=0.2$ balances accuracy (78.58\%) and stability (100\% convergence), validating convergence theorem's bound $d(\theta^*, \theta_{\text{Fr}}) \leq \frac{\lambda}{1-\lambda}d(\theta_{\text{Lor}}, \theta_{\text{Fr}})$. % RLPF performance across varying regularization strength $\lambda \in [0.01, 1.0]$ and number of optimization steps. Method shows robust performance with $\lambda=0.2$ achieving optimal accuracy-stability trade-off. Convergence typically occurs within 3-5 iterations. 

\subsection{Fusion controls and optimization stages}
\label{sec:fusion_controls}

\begin{figure}[t]
\centering
\includegraphics[width=\linewidth]{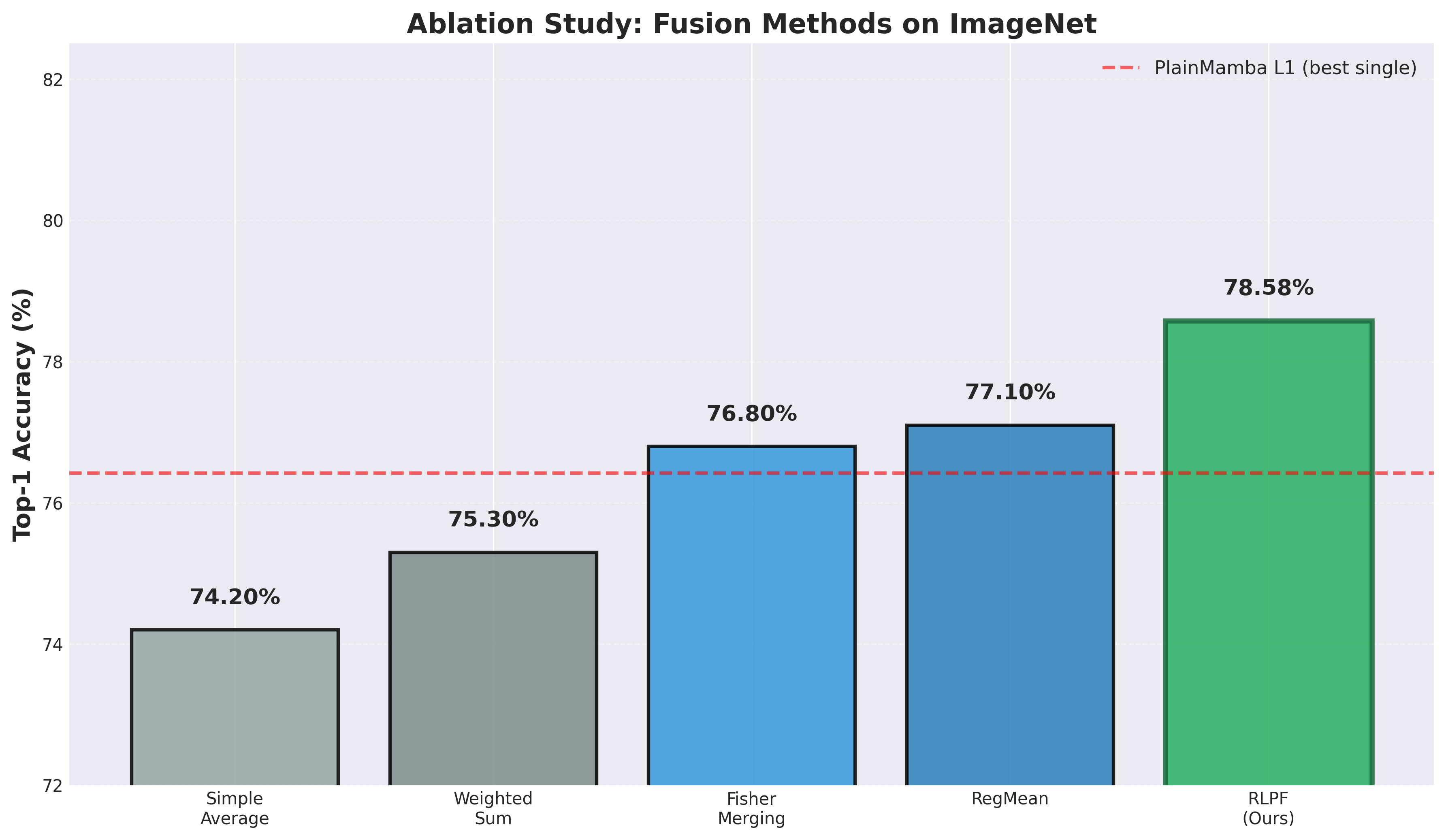}
\caption{Available ImageNet fusion-method comparison. The figure records 74.20\% for simple averaging, 75.30\% for a weighted sum, 76.80\% for Fisher merging, 77.10\% for RegMean, and 78.58\% for RLPF. A controlled interpretation requires confirmation that all methods used the same alignment, gate, and post-fusion optimization budget.}
\label{fig:fusion_methods}
\end{figure}

\paragraph{Fusion-method comparison.}
Figure~\ref{fig:fusion_methods} records an ImageNet comparison among simple averaging (74.20\%), a weighted sum (75.30\%), Fisher merging (76.80\%), RegMean (77.10\%), and RLPF (78.58\%). All four alternatives fall below RLPF in this figure, and simple averaging also falls below the 76.42\% best-parent reference. This is useful evidence that the reported result is not reproduced by these alternative fusion rules. It is not yet a fully controlled comparison, however: the figure and accompanying report do not identify the alignment map, post-fusion optimizer, number of epochs, gate, or checkpoint-selection rule for each method. We therefore report the observed ordering without attributing the difference solely to Lorentz geometry.

\paragraph{Initialization versus full fine-tuning.}
The available ImageNet record provides two RLPF stages: 77.80\% immediately after fusion initialization and 78.58\% after the recorded fine-tuning, a gain of 0.78 percentage points. Both values exceed the 76.42\% best-parent point estimate. This comparison suggests that the initialization contributes substantially to the final result, while supervised optimization supplies an additional gain. No intermediate few-epoch checkpoints are recorded, so a zero/few/full fine-tuning curve cannot be reconstructed from the available artifacts.

\paragraph{What the convergence figures establish.}
The files \emph{learning\_curves.png} and \emph{convergence\_analysis.png} visualize training loss, validation accuracy, and epochs required to reach fractions of final accuracy. They concern optimization dynamics, not numerical failures of Poincar\'e, Lorentz, or Riemannian fusion. They therefore cannot estimate a failure probability or support a claim of guaranteed numerical stability. Such a test requires repeated merge attempts and a pre-specified failure event, such as any NaN/Inf, violation of the manifold constraint beyond a tolerance, or failure to reach a gradient/residual threshold within a fixed iteration budget.

\subsection{Efficiency, robustness, and reproducibility}
\label{sec:limitations}

\paragraph{Available resource accounting.}
Table~\ref{tab:resource_accounting} transcribes the values displayed in the supplied ImageNet efficiency artifact. These values are preferable to borrowing DeiT or VMamba measurements because they refer to the model names used in this study. They should nevertheless be treated as provisional: the artifact does not record hardware, batch size, image resolution, numerical precision, FLOP-counting convention, warm-up, number of timing repetitions, or whether RLPF evaluates both branches. The displayed model sizes also differ from the 5.7M/7.3M CIFAR-10 parameter counts recorded above, indicating a different configuration or an unresolved metadata inconsistency.

\begin{table}[t]
\centering
\caption{Resource values displayed in the available ImageNet efficiency artifact. Latency is reported as mean $\pm$ the plotted error bar; the statistic represented by that error bar is not recorded. Training cost was not provided.}
\label{tab:resource_accounting}
\setlength{\tabcolsep}{3.2pt}
\vspace{-0.9em}
\begin{tabular}{lcccc}
\toprule
Model & Params & FLOPs & Memory & Latency \\
 & (M) & (G) & (MB) & (ms) \\
\midrule
ViT/Transformer & 22.20 & 4.60 & 450 & $18.5\pm0.8$ \\
PlainMamba/SSM & 7.00 & 3.00 & 380 & $15.2\pm0.6$ \\
RLPF hybrid & 22.50 & 4.78 & 520 & $20.1\pm0.9$ \\
\bottomrule
\end{tabular}
\end{table}

Under the displayed measurements, PlainMamba has the lowest parameter count, FLOPs, memory, and latency, whereas RLPF has the highest values and the highest recorded ImageNet accuracy. RLPF is approximately 8.6\% slower than the displayed ViT latency and 32.2\% slower than PlainMamba, so these data support an accuracy--resource trade-off rather than an inference-speedup claim. A valid final comparison requires one fixed protocol covering hardware, software versions, batch size, precision, image resolution, warm-up, timing repetitions, and gate policy. If soft routing evaluates both branches before logit fusion, it cannot realize single-branch latency without a separately defined hard-routing mechanism.

\section{Conclusion}
\label{sec:conclusion}

RLPF is a framework for initializing and training a gated ViT--SSM hybrid from heterogeneous parents. It uses role-based latent alignment, component-dependent Euclidean or Lorentz fusion, and a geodesic anchor objective. The recorded point estimates exceed the best parent on three image-classification datasets, with the strongest documented ImageNet result increasing from 77.80\% before fine-tuning to 78.58\% after fine-tuning. These observations are promising but do not yet isolate geometry from alignment, routing, or additional optimization. The method is not a training-free single-checkpoint merge, and the available evidence does not support claims of unconditional numerical stability, inference acceleration, statistical significance, or billion-parameter scalability. Addressing the controlled-baseline and reproducibility gaps above is necessary for a defensible evaluation.

{
    \small
    \bibliographystyle{ieeenat_fullname}
    \bibliography{main}
}
\appendix
\section{Supplementary Material}
\label{sec:supplement}

This supplement provides mathematical checks, implementation-level definitions, and the experimental settings available for Riemannian--Lorentz Parameter Fusion (RLPF). It uses the same notation as the main paper and does not introduce additional performance claims.

\subsection{Lorentz construction}
\label{sec:supp_lorentz}

For curvature magnitude $c>0$, the Lorentz hyperboloid is
\begin{equation}
 \mathbb{H}^{r}_{c}=\left\{h\in\mathbb{R}^{r+1}:\langle h,h\rangle_L=-1/c,\ h_0>0\right\},
 \label{eq:supp_hyperboloid}
\end{equation}
where $\langle h,k\rangle_L=-h_0k_0+\sum_{j=1}^{r}h_jk_j$. The lift used by RLPF maps $q\in\mathbb{R}^{r}$ to
\begin{equation}
 E_c(q)=\left(\sqrt{c^{-1}+\lVert q\rVert_2^2},q\right).
 \label{eq:supp_lift}
\end{equation}
Substitution verifies that
\begin{equation}
 \langle E_c(q),E_c(q)\rangle_L
 =-\left(c^{-1}+\lVert q\rVert_2^2\right)+\lVert q\rVert_2^2=-1/c,
\end{equation}
so the lifted point satisfies Eq.~\eqref{eq:supp_hyperboloid}.

For $p,q\in\mathbb{H}^{r}_{c}$, define $a=-c\langle p,q\rangle_L\geq1$. The geodesic distance and logarithmic map are
\begin{align}
 d_c(p,q)&=\frac{1}{\sqrt{c}}\operatorname{arcosh}(a),
 \label{eq:supp_distance}\\
 \log_p(q)&=\frac{\operatorname{arcosh}(a)}{\sqrt{a^2-1}}(q-ap),
 \label{eq:supp_log}
\end{align}
with the continuous limit used when $p=q$. For a tangent vector $v\in T_p\mathbb{H}^{r}_{c}$, where $\langle p,v\rangle_L=0$, the exponential map is
\begin{equation}
 \exp_p(v)=\cosh(\sqrt{c}\lVert v\rVert_L)p
 +\frac{\sinh(\sqrt{c}\lVert v\rVert_L)}{\sqrt{c}\lVert v\rVert_L}v,
 \label{eq:supp_exp}
\end{equation}
where $\lVert v\rVert_L=\sqrt{\langle v,v\rangle_L}$ on the tangent space. The ratio in Eq.~\eqref{eq:supp_exp} is evaluated by its limit at $v=0$~\cite{law2019lorentzian}.

\begin{figure*}[t]
\centering
\includegraphics[width=\textwidth]{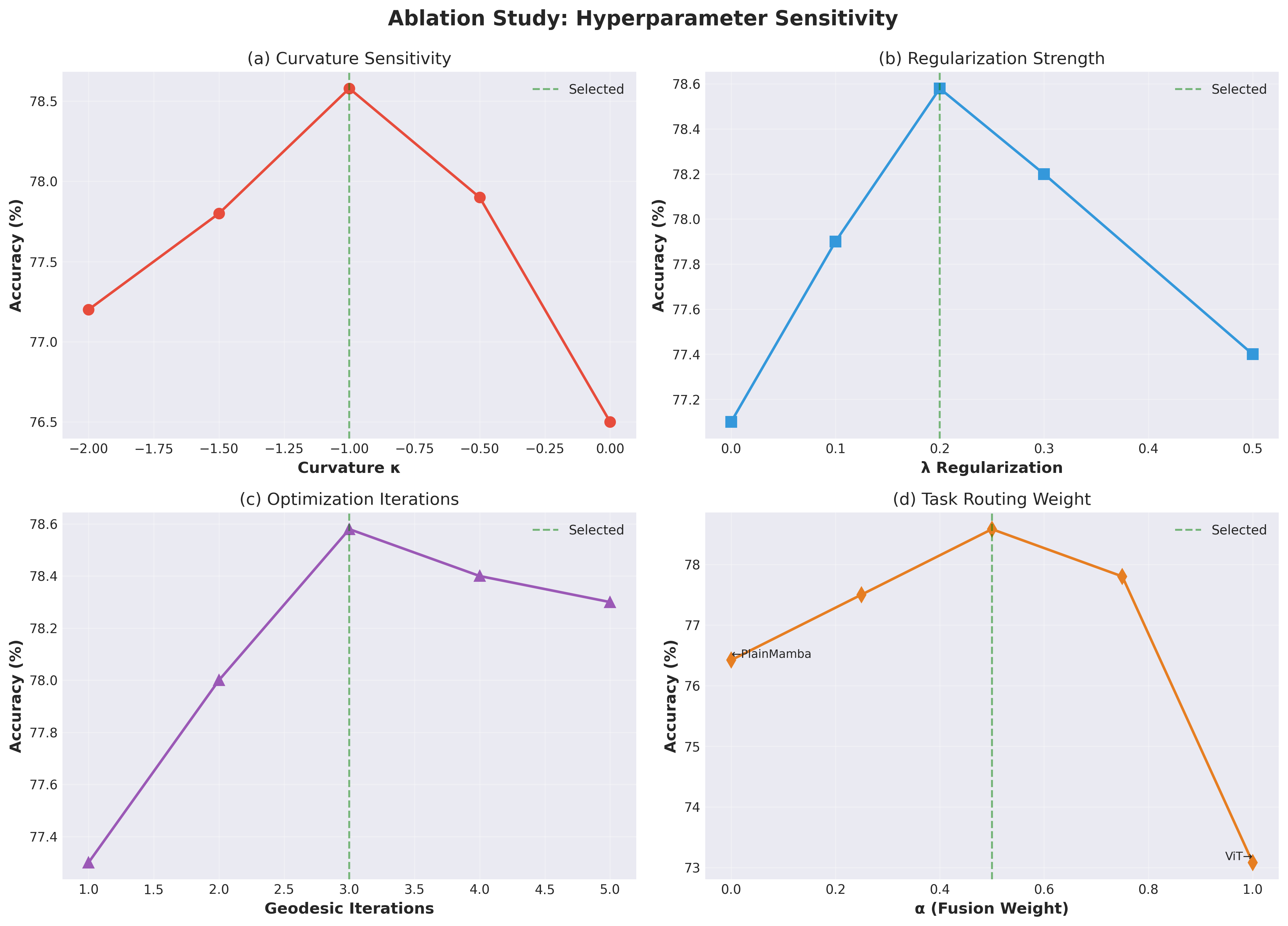}
\caption{Available ImageNet-1K hyperparameter sweeps. Dashed lines mark the selected settings in the artifact. Point estimates are shown without uncertainty; the figure therefore documents sensitivity of the recorded run rather than statistical robustness.}
\label{fig:supp_hyperparameters}
\end{figure*}

\subsection{Anchor and objective}
\label{sec:supp_objective}

Let $h_{\mathrm{V},g},h_{\mathrm{S},g}\in\mathbb{H}^{r_g}_{c_g}$ denote the lifted ViT and SSM representations for component group $g$, and let $w_{\mathrm{V}}+w_{\mathrm{S}}=1$ with nonnegative weights. Define
\begin{equation}
 \bar h_g=w_{\mathrm{V}}h_{\mathrm{V},g}+w_{\mathrm{S}}h_{\mathrm{S},g},
 \qquad
 h_{0,g}=\frac{\bar h_g}{\sqrt{-c_g\langle\bar h_g,\bar h_g\rangle_L}}.
 \label{eq:supp_anchor}
\end{equation}
When the denominator is real and nonzero, direct substitution gives $\langle h_{0,g},h_{0,g}\rangle_L=-1/c_g$. The positive time component is retained, placing $h_{0,g}$ on the upper sheet of the hyperboloid.

The RLPF objective for group $g$ is
\begin{equation}
 \boxed{\begin{aligned}
 h_g^*=\arg\min_{h\in\mathbb{H}^{r_g}_{c_g}}\ \Bigl\{
 &\underbrace{\sum_{a\in\{\mathrm{V},\mathrm{S}\}}w_a d_{c_g}^2(h,h_{a,g})}_{\text{geodesic fidelity}}\\[-1mm]
 &+\lambda\underbrace{d_{c_g}^2(h,h_{0,g})}_{\text{anchor penalty}}\Bigr\}.
 \end{aligned}}
 \label{eq:supp_objective}
\end{equation}
The regularizer is a squared geodesic distance. An ambient quantity such as $\langle h-h_0,h-h_0\rangle_L$ is not used because the Minkowski inner product is indefinite and therefore does not define a nonnegative norm.

Hyperbolic space is a Hadamard manifold. Consequently, a positive weighted sum of squared distances is geodesically convex and has a unique minimizer~\cite{karcher1977riemannian,pennec2006intrinsic}. This establishes existence and uniqueness of the exact minimizer of Eq.~\eqref{eq:supp_objective}; it does not establish a finite-iteration rate for the step schedule used in the implementation.

Starting from $h_g^{(0)}=h_{0,g}$, RLPF uses
\begin{align}
 v_g^{(t)}&=\sum_a w_a\log_{h_g^{(t)}}(h_{a,g})+\lambda\log_{h_g^{(t)}}(h_{0,g}),\\
 h_g^{(t+1)}&=\exp_{h_g^{(t)}}\left(\alpha_t v_g^{(t)}\right),
 \qquad \alpha_t=\frac{1}{1+t}.
 \label{eq:supp_update}
\end{align}
The reported configuration applies three updates. This is a fixed computational budget, not a claim that every instance reaches the exact minimizer in three iterations.

\subsection{Numerical implementation}
\label{sec:supp_numerics}

The hyperboloid has no finite-radius boundary, but finite-precision operations can still overflow or drift from the manifold constraint. A reliable implementation should:
\begin{enumerate}
 \item compute Lorentz inner products in sufficient precision;
 \item clamp the argument of $\operatorname{arcosh}$ to at least $1+\varepsilon$;
 \item evaluate $\operatorname{arcosh}(a)/\sqrt{a^2-1}$ and $\sinh(x)/x$ with stable small-argument limits;
 \item verify $\langle h,h\rangle_L\approx-1/c$ after each update and reproject when needed;
 \item record explicit tolerances for constraint violation, non-finite values, and non-convergence.
\end{enumerate}
These safeguards reduce numerical risk but do not provide an unconditional guarantee against NaN or overflow. A numerical-stability rate requires repeated trials and a pre-specified failure criterion.
\begin{table*}[t]
\centering
\caption{Training information available for the three reported datasets.}
\label{tab:supp_protocol}
\setlength{\tabcolsep}{5pt}
\begin{tabular}{lccc}
\toprule
Setting & CIFAR-10 & Oxford-IIIT Pet & ImageNet-1K \\
\midrule
Parent families & ViT-Tiny / PlainMamba-L1 & same families & same families \\
Initialization & from scratch & ImageNet pretrained & pretrained; source unavailable \\
Recorded epochs & 50 & 50 & 50; stage separation unavailable \\
Optimizer & SGD & AdamW & SGD \\
Learning rate & $0.1$ & $10^{-4}$ & $0.1$ \\
Batch size & 1024 & unavailable & 1024 \\
Weight decay / schedule & unavailable & unavailable & unavailable \\
Augmentation / preprocessing & unavailable & unavailable & unavailable \\
Checkpoint-selection rule & unavailable & unavailable & unavailable \\
Reported metric & accuracy & accuracy & top-1 accuracy \\
\bottomrule
\end{tabular}
\end{table*}

\subsection{Cross-architecture alignment}
\label{sec:supp_alignment}

RLPF pairs components by computational role rather than asserting equality between their operators. Table~\ref{tab:supp_alignment} records the correspondence used in the method.

\begin{table}[t]
\centering
\caption{Role-level correspondence between the retained ViT and SSM branches.}
\label{tab:supp_alignment}
\setlength{\tabcolsep}{3pt}
\begin{tabular}{lll}
\toprule
Role & ViT parameters & SSM parameters \\
\midrule
Embedding & patch embedding & patch embedding \\
Sequence mixing & $W_Q,W_K,W_V,W_O$ & $A,B,C,\Delta$ \\
Pointwise map & feed-forward layers & feed-forward layers \\
Normalization & scale and shift & scale and shift \\
\bottomrule
\end{tabular}
\end{table}

For architecture $a\in\{\mathrm{V},\mathrm{S}\}$ and group $g$, the parameters are serialized as $u_{a,g}=\operatorname{vec}(\theta_{a,g})$ and projected to $q_{a,g}=P_{a,g}u_{a,g}\in\mathbb{R}^{r_g}$. Fusion is performed in the shared dimension $r_g$, after which $P_{a,g}^{\mathsf T}q_g^*$ is reshaped for the corresponding branch. Shape-compatible groups may use identity maps.

A complete implementation must fix the tensor serialization order, construction of each SVD input matrix, rank $r_g$, singular-vector sign convention, handling of discarded directions, decoder, and unmatched parameters. These choices are not determined by the phrase ``truncated SVD'' and should be distributed with the implementation. Useful controls include random orthogonal maps at the same rank, Euclidean fusion through the same maps, identity maps for compatible groups, and learned adapters with matched training budgets.

\subsection{Component settings}
\label{sec:supp_components}

Table~\ref{tab:supp_curvature} lists the component settings used in the reported experiments. The values are design hyperparameters selected for the aligned latent representation; they are not measurements of the intrinsic curvature of the original network weights.

\begin{table}[t]
\centering
\caption{Recorded component-wise curvature magnitudes. Curvature is $-c_g$ when $c_g>0$.}
\label{tab:supp_curvature}
\setlength{\tabcolsep}{4pt}
\begin{tabular}{lcc}
\toprule
Group $g$ & $c_g$ & Fusion rule \\
\midrule
Embedding & $1.0$ & Lorentz \\
Sequence mixing & $0.1$ & Lorentz \\
Feed-forward & $0.001$ & near-Euclidean Lorentz \\
Normalization & $0$ & Euclidean \\
\bottomrule
\end{tabular}
\end{table}

The other recorded settings are anchor weight $\lambda=0.2$, number of intrinsic updates $T=3$, and gate regularization $\tau=0.01$. The gate penalty $\tau(\beta_\phi(x)-0.5)^2$ favors balanced branch weights; it is not a diversity objective.

\subsection{Additional ablation figures}
\label{sec:supp_ablations}

\begin{figure}[t]
\centering
\includegraphics[width=\linewidth]{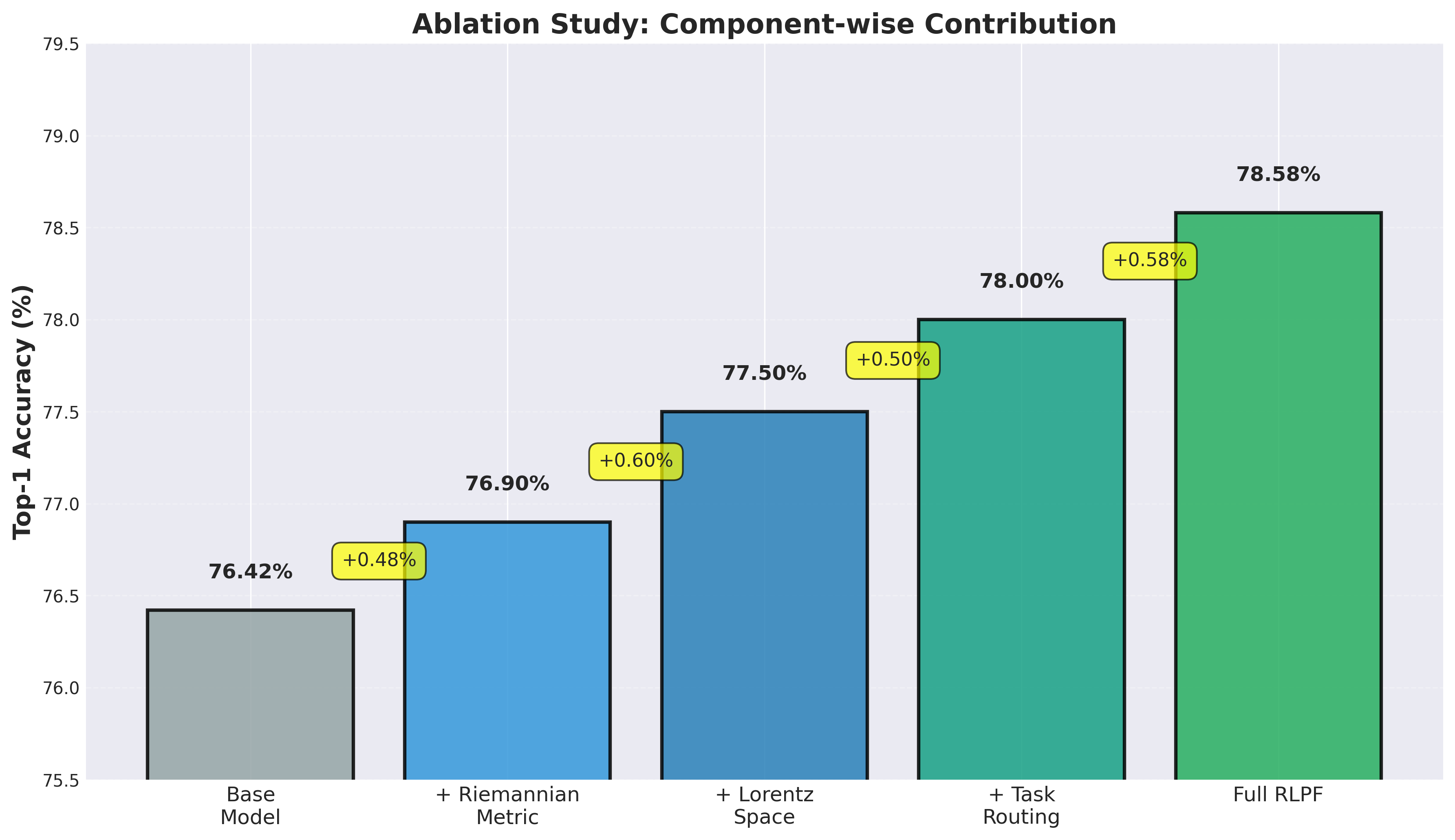}
\caption{Available cumulative component ablation on ImageNet-1K. The sequence of configurations ends at the reported RLPF result of 78.58\%. Because each bar adds a component to the preceding configuration, the increments do not constitute one-at-a-time removal effects.}
\label{fig:supp_component_ablation}
\end{figure}

\paragraph{Incremental component ablation.}
Figure~\ref{fig:supp_component_ablation} records the available ImageNet component study. Starting from the 76.42\% best-parent reference, the plotted configurations obtain 76.90\% after adding the Riemannian metric, 77.50\% after adding Lorentz-space fusion, 78.00\% after adding task routing, and 78.58\% for the complete configuration. These are cumulative configurations, so the differences should not be interpreted as independent causal effects of individual components. The artifact contains no repeated-trial uncertainty.

\paragraph{Hyperparameter sensitivity.}
Figure~\ref{fig:supp_hyperparameters} reports the available sweeps over curvature, anchor weight $\lambda$, number of intrinsic updates, and a fixed fusion weight. The selected settings shown in the artifact are curvature $-1$, $\lambda=0.2$, and $T=3$. The curves support local sensitivity analysis around the selected configuration, but no error bars, tuning split, or repeated runs are available. The fourth panel concerns a fixed interpolation weight and is distinct from the input-dependent gate $\beta_\phi(x)$ in Eq.~\eqref{eq:routing}.

\subsection{Available training protocol}
\label{sec:supp_protocol}

Table~\ref{tab:supp_protocol} contains only settings present in the manuscript records. Unavailable fields are stated explicitly rather than inferred.

\end{document}